\documentclass[conference]{IEEEtran}
\usepackage{cite}
\usepackage{graphicx}
\usepackage{url}
\usepackage{amsmath}
\usepackage{amssymb}
\usepackage[caption = false, font=footnotesize]{subfig}
\usepackage{float}
\usepackage{booktabs} 
\usepackage{multirow}
\usepackage[colorlinks=true,linkcolor=blue]{hyperref}%
\usepackage{array}
\usepackage{color}
\usepackage{tabularx}
\usepackage{longtable}
\usepackage{tabu}
\usepackage{pifont}
\usepackage{amssymb}
\usepackage{ragged2e}
\usepackage{tabularx}
\usepackage{longtable}
\usepackage{booktabs}

\ifCLASSINFOpdf
\else
\fi
\begin{document}
%
\title{Generative Visual Compression: A Review}



%
\author{\IEEEauthorblockN{Bolin Chen\IEEEauthorrefmark{1},
Shanzhi Yin\IEEEauthorrefmark{1},
Peilin Chen\IEEEauthorrefmark{1}, 
Shiqi Wang\IEEEauthorrefmark{1} and
Yan Ye\IEEEauthorrefmark{2}}
\IEEEauthorblockA{\IEEEauthorrefmark{1} City University of Hong Kong} 
 \IEEEauthorrefmark{2} Alibaba Group}


\maketitle

\begin{abstract}
Artificial Intelligence Generated Content (AIGC) is leading a new technical revolution for the acquisition of digital content and impelling the progress of visual compression towards competitive performance gains and diverse functionalities over traditional codecs. This paper provides a thorough review on the recent advances of generative visual compression, illustrating great potentials and promising applications in ultra-low bitrate communication, user-specified reconstruction/filtering, and intelligent machine analysis. In particular, we review the visual data compression methodologies with deep generative models, and summarize how compact representation and high-fidelity reconstruction could be actualized via generative techniques. In addition, we generalize related generative compression technologies for machine vision and intelligent analytics. Finally, we discuss the fundamental challenges on generative visual compression techniques and envision their future research directions.
\end{abstract}

%

%
\IEEEpeerreviewmaketitle

\section{Introduction}
The concept ``generative visual compression" was first mentioned in~\cite{8456298}, which utilizes deep generative models (\textit{i.e.,} Variational Auto-Encoder (VAE)~\cite{VAE}, Generative Adversarial Network (GAN)~\cite{goodfellow2014generative} and Diffusion Model (DM)~\cite{ho2020denoising}) to compress the visual data in pursuit of realizing visually-pleasing reconstructions within the minimal coding costs compared with traditional image/video compression algorithms. Especially in the past two years, the cumulative integration of generative models, Contrastive Language-Image Pre-Training (CLIP)~\cite{radford2021learning}, Transformer~\cite{vaswani2017attention} and other technologies has given rise to the explosion of Artificial Intelligence Generated Content (AIGC), making it more versatile in digital content creation.

Essentially, different from discriminative models predicting the decision boundary between the classes accompanied by insensitivity to outliers, deep generative models can well allow for data generation and augmentation since they focus on  learning the underlying patterns or distribution from a given set of data. Similarly, visual data compression also aims to establish the relationship between an index and an image (a point) in the high dimensional image space, such that it can exploit statistical redundancy to represent data or utilize strong information prior to decompose/encode signal towards optimal rate-distortion trade-offs. As such, there are clear commonalities between the generation and compression tasks, providing great possibilities for exploring generative visual compression. Specifically, deep generative models are able to learn the compact feature distribution required in compression tasks, whilst their strong inference capabilities also facilitate the signal reconstruction from these compact distributions. Different from traditional hybrid coding frameworks such as H.264/Advanced Video Coding (AVC)~\cite{wiegand2003overview}, H.265/High Efficiency Video Coding (HEVC)~\cite{sullivan2012overview} and H.266/
Versatile Video Coding (VVC)~\cite{bross2021overview}, these novel generative compression paradigms possess more compact feature representations, flexible motion estimation mechanisms, and superior signal reconstruction capabilities, which can bring promising compression performance and diverse functionalities.

This paper provides a review on generative visual compression for both human and machine visions. In particular, we generalize a wide spectrum of generative visual compression methodologies and explore the inner correlations between generation and compression. Furthermore, we introduce the technical evolution from human vision to machine vision, aiming at constructing more intelligent coding and collaborative analysis systems. Finally, we summarize basic challenges and envision future possible research for these state-of-the-art generative visual compression schemes.

\section{Generative Visual Compression for Human Vision}
\begin{figure*}[tb]
\centering
\includegraphics[width=0.96\textwidth]{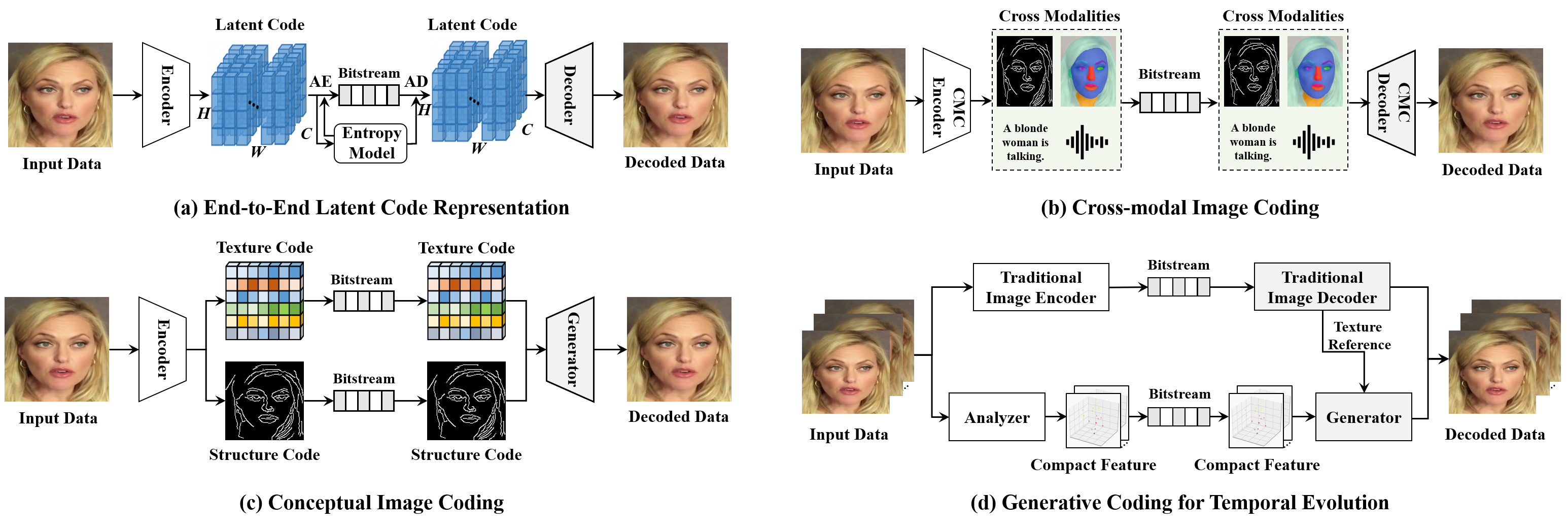}
\caption{Illustration of typical generative compression frameworks for human vision.} 
\label{fig1} 
\end{figure*}

The majority of current image/video compression research aims at improving the visually-pleasing experience for the human visual system within minimal coding bit rates. This section will review the progress of existing generative compression algorithms to clarify how well deep generative models can perform in visual compression tasks, mainly including end-to-end latent code representation, cross-modal image coding, conceptual image coding, generative coding for temporal evolution, and omni-dimensional data coding as illustrated in Fig.~\ref{fig1}.

\subsection{End-to-End Latent Code Representation}
Benefiting from the strong inference capacity of deep generative networks, learned image/video coding algorithms can jointly optimize the entire encoder/decoder towards the rate-distortion trade-off in an end-to-end trainable manner. Ball{\'e} \textit{et al.}~\cite{balle2016end} firstly explored the rate–distortion optimization problem within the context of VAE and proposed an end-to-end learned image compression framework, where natural image can be mapped to a latent code space via a parametric analysis transform. Following this, scale hyper-priors~\cite{balle2018variational}, autoregressive context model~\cite{NEURIPS2018_53edebc5}, Gaussian mixture model~\cite{Cheng_2020_CVPR}, 
channel-wise model\cite{minnen2020channel}, 
channel-spatial model\cite{he2022elic} 
were successively employed to parameterize the distributions of latent codes and optimize the entropy model, achieving competitive compression performance compared to traditional codecs.
In addition, the rate-distortion-complexity optimization~\cite{zhang2023elfic,yang2021slimmable} was further conducted such that they can support variable rate and flexible decoding complexity. Besides, adversarial learning~\cite{Agustsson_2019_ICCV,NEURIPS2020_8a50bae2} was also introduced to
improve visual quality. The recent progress of diffusion models also improved the performance of learned image compression with residual augmentation~\cite{goose2023neural} and deterministic inference~\cite{yang2022lossy}.

As for learned video compression, Habibian \textit{et al.}~\cite{Habibian_2019_ICCV} introduced a rate-distortion auto-encoder, and Lombardo \textit{et al.}~\cite{NEURIPS2019_f1ea154c} employed a sequential VAE to exploit spatiotemporal redundancy based upon learned image codec framework. In addition, Li \textit{et al.}~\cite{Li_2023_CVPR} utilized context diversity in both temporal and spatial dimensions to enhance the neural video codec that could surpass the under-development next-generation traditional codec ECM. Furthermore, GAN-based neural video compression methods~\cite{mentzer2021neural,zhao_DPEG,Liu_2021_CVPR} have been developed for realistic synthesis and redundancy reduction.

\subsection{Cross-modal Image Coding}
Multi-modality data (\textit{e.g.,} auditory, textual, and haptic) have been widely used in various vision tasks for robust perception and understanding. Relying on such cross-modal learning~\cite{8269806}, visual data compression can be further developed towards high-level human-comprehensible communication. In particular, Li \textit{et al.}~\cite{Liacmmm} proposed the first cross-modal compression framework that can encode images into compact text for semantic communication. Besides, Zhang \textit{et al.}~\cite{10032603} followed the philosophy of scalable coding to design a scalable cross-modality compression paradigm with different modalities. To improve the semantic fidelity, Gao \textit{et al.}~\cite{10125473} developed a novel rate-distortion optimized cross-modal coding scheme using reinforcement learning. In addition, a variable-rate cross-modal compression mechanism~\cite{10236483} was proposed to meet the demands of changeable transmission bandwidth. Tandon \textit{et al.}~\cite{9953071} presented a compression pipeline that can convert visual data into text transcript for dramatically reducing data transmission rates, whilst Lu \textit{et al.}~\cite{Lu_2022_CVPR} employed different visual data types (\textit{i.e.,} infrared and visible image pairs) to exploit the cross-modal redundancy.

\subsection{Conceptual Image Coding}
Inspired by visual decomposition tasks~\cite{jeon2014intrinsic,kim2018structure}, conceptual coding is proposed to disentangle natural images into a series of conceptual representations for high-efficiency compression. In details, the conceptual compression framework firstly encodes visual image data into structure information and texture code, and then achieves image decoding via a deep generative model. Following this paradigm, Chang \textit{et al.}~\cite{chang2019layered} chose to compactly encode critical structural edge and color/texture information, which can achieve promising performance over traditional image codecs at similar bitrate. On this basis, semantic prior modeling~\cite{chang2022conceptual}, consistency-contrast regularization~\cite{chang2022consistency} and semantic-aware visual decomposition~\cite{chang2023semantic} were further proposed to preserve higher coding flexibility and better visual reconstruction at extreme bitrates.

\subsection{Generative Coding for Temporal Evolution}
Current generative video coding algorithms mainly exploit strong priors or learn temporal dynamics between video frames to formulate the analysis-synthesis-based compression framework~\cite{chen2023generative}, thus achieving low-bandwidth communication. Specifically, the encoder employs traditional coding technique like HEVC or VVC to compress key-reference frames of video sequence and an analysis model to characterize subsequent inter frames into compact transmitted symbols, whilst the decoder can utilize the deep generative model to synthesize video frames from the decoded key-reference frames and compact temporal information. 

Taking human face/body videos as examples, they possess much inherent structure and prior knowledge, such as their shape, composition, and movement, which can be easily compressed with this methodology. Based on First Order Motion Model (FOMM)~\cite{siarohin2019first}, an end-to-end animation model, the generative video compression framework can be optimized towards ultra-low bit-rate communication~\cite{facebook2021,ultralow,9859867}, where video frames are characterized into a series of learned 2D keypoints. Besides, Oquab \textit{et al.}~\cite{facebook2021} developed the first real-time generative compression system on the mobile platform via the SPADE architecture and segmentation maps. In addition, some facial priors like 2D landmarks~\cite{feng2021generative}, 3D semantics~\cite{chen2023interactive}, and compact feature representation ~\cite{CHEN2022DCC} are employed as the transmitted animation symbols. Wang \textit{et al.}~\cite{10181664} also proposed an ultra-low bitrate video codec via a PCA-based decomposing method, greatly allowing for the compression potential of motion representations. Regarding human body video compression, a disentangled texture-structure visual representation~\cite{9859831} was proposed, where human pose keypoints are leveraged as the structure code to achieve extremely low bitrate. Moreover, to increase compression performance and broaden application scenarios, various novel technologies have been applied to this generative video compression framework, such as frame interpolation~\cite{compressing2022bmvc}, residual enhanced coding~\cite{konuko2023predictive}, spatial-temporal adversarial training~\cite{chen2023csvt}, multi-view aggregation~\cite{volokitin2022neural}, multi-reference dynamic prediction~\cite{icip2022zhao} and feature transcoding~\cite{yin2024parametertranslator}. Indeed, such low-bitrate generative video compression paradigm is entirely possible to be applied to other natural scenes~\cite{li2023generative} with oscillatory dynamics and motion prior, such as trees, flowers, candles, and clothes swaying in the wind.

\subsection{Omni-Dimensional Data Coding}
Unlike visual data from natural scenes, omni-dimensional signals like 3D point clouds, light fields, and 360-degree data can represent both static and dynamic objects/scenarios with a higher number of dimensions, thus offering the advantage of a more realistic and immersive visual experience. Indeed, this also means that the compression/transmission of such raw omni-dimensional data requires more coding bits. Thus, expanding generative visual compression algorithms into the high dimensional signal data may be one solution to achieving their compact representation and efficient reconstruction. 

In particular, for the compression of point cloud geometry, Wang \textit{et al.}~\cite{wang2021lossy} proposed a novel end-to-end VAE-based framework to capture compact latent features and actualize hyperprior generation.
In addition, He \textit{et al.}~\cite{he2022density} designed an auto-encoder to preserve local density information, and Nguyen \textit{et al.}~\cite{nguyen2021lossless} employed a deep generative model to estimate the probability distribution of voxel occupancy. As for light field compression, Jia \textit{et al.}~\cite{jia2018light} and Liu \textit{et al.}~\cite{liu2021view} both proposed GAN-based view synthesis-compression methods with a cascaded hierarchical coding structure and quality enhancement technique, respectively. Also, quantization-aware learning~\cite{jiang2022untrained} and dual discrimination models~\cite{bakir2020light} are utilized in light field compression tasks for promising performance. Moreover, generative compression in other high dimensional data scenes, such as stereo image~\cite{liu2019dsic} and 360-degree data~\cite{li2022end}, have also been explored for visual perception applications.

\begin{figure*}[tb]
\centering
\includegraphics[width=0.96\textwidth,height=6cm]{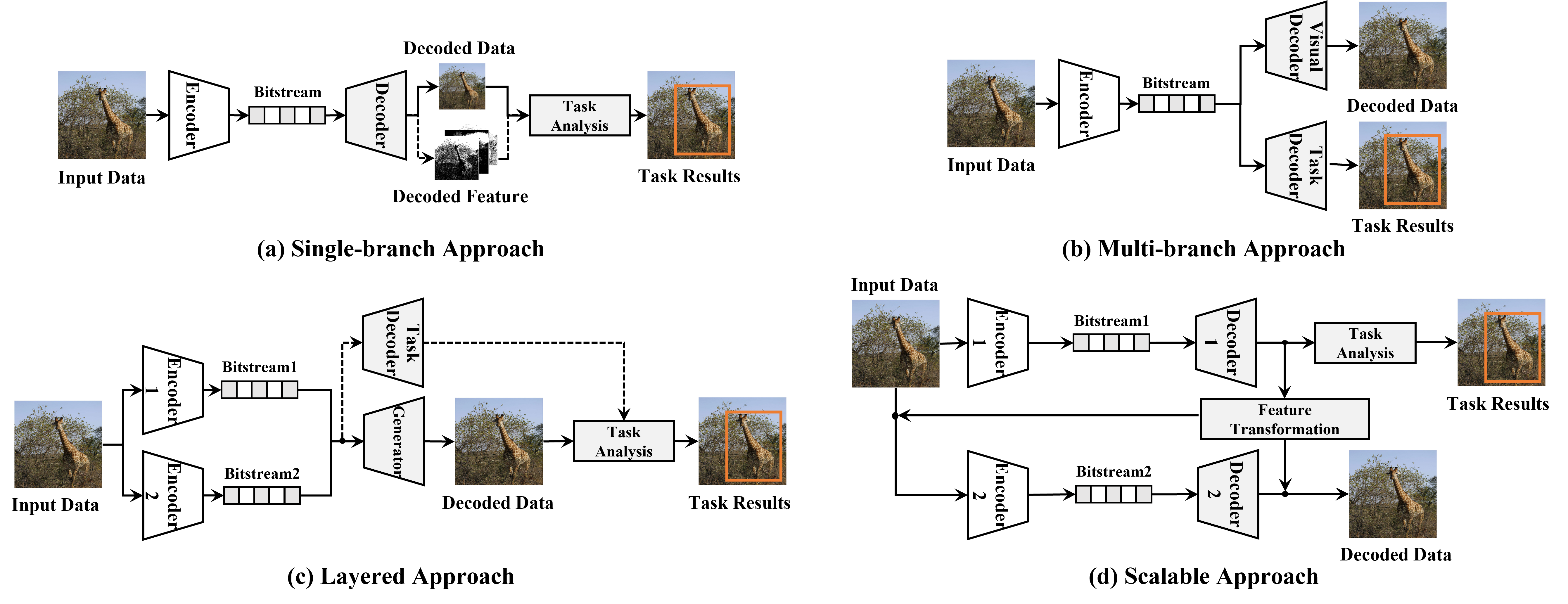}
\caption{Illustration of typical generative compression frameworks for machine vision. Dash lines show the optional structures.} 
\label{fig2} 
\end{figure*} %

\section{Generative Visual Compression for Machine Vision}
With a considerable amount of visual data being ultimately received and processed by machine for high-level vision tasks, visual compression for machine vision has been widely studied in recent years\cite{Duan_2022}. It aims to maintain machine task performance from compressed visual data. Similar to human-oriented scenarios, many research works leverage the generative model as the backbone of machine-oriented visual compression networks and implement end-to-end optimization by combining rate, distortion, and machine performance. 

In this section, we first categorized these methods based on different domains that general machine tasks are performed on. Specifically, if the machine task is performed via reconstructed pixel-level image/video, it is categorized as ``Pixel Domain Analysis". On the other hand, if the machine analysis is performed directly on transformed generative feature, we categorized such methods as ``Feature Domain Analysis". In addition, regardless of the requirement for machine performance, pixel-level reconstructions are available as an intermediate step for machine analysis or additional tasks in most cases. We further categorize these methods based on their arrangement for human and machine vision as ``Single-branch", ``Layered", ``Multi-branch" or ``Scalable". The illustration of those different pipelines is shown in Fig.~\ref{fig2}.

\subsection{Pixel Domain Analysis}
\subsubsection{Single-branch Approaches}
The single-branch approaches of generative compression for machine vision are explored by directly feeding reconstructed visual data from learned visual data codec to machine task networks. Such works mainly focus on optimization strategies or compression network design. A study regarding end-to-end image compression for machine vision was conducted in~\cite{e2e4m_study_DCC2022} by comparing different training and tuning methods with off-the-shelf task and compression models. Le \textit{et al.}~\cite{Le_weighting} proposed a weighting schedule to balance the trade-offs among rate, distortion and task performance during training. Also, they proposed an online inference-time fine-tuning strategy to update latent features by considering the feature distortion of the task network as a proxy loss~\cite{Le_infertune}. In addition, Wang \textit{et al.}~\cite{wang2022end} designed a unified optimization framework with generalized rate-accuracy optimization and variable bit-rate coding. They also explored channel number distribution for an inverted bottleneck encoder structure~\cite{wang2021end}. To realize machine-friendly reconstruction and low bit-rate compression for underwater images, feature enhancement with prior-guided contrastive learning~\cite{underwater2} was leveraged to improve decoder-side feature preservation and alleviate underwater degradation.

Efforts have also been made to expand the data dimension or task diversities. Yi \textit{et al.}~\cite{Yi2023} explored the framework design and optimization such that end-to-end video compression can also achieve machine analysis, where multi-scale motion estimation and multi-frame feature fusion were proposed with task-driven optimization to trade-off between signal-level and semantic-level fidelity. To utilize multiple machine tasks, the connectors were proposed to adapt reconstructions from a primary machine task to multiple secondary tasks~\cite{chamain2021end}. Similarly, Chen \textit{et al.}~\cite{TransIC2023} transferred a well-trained transformer-based image codec to various machine tasks by injecting instance-specific prompts and task-specific prompts to the encoder and decoder.

\subsubsection{Layered/Scalable Approaches}
While single-branch methods are intuitive and convenient for implementation, such a naive combination of compression and machine task model lacks flexibility and hinders further performance improvement. Inspired by conceptual coding in human-oriented vision compression, there have been latest works realizing layered structure with multiple bit-streams or scalable structures with multi-level reconstructions. In the meantime, most of them take advantage of GANs to obtain ultra-low bit-rate and semantically high quality with dedicated feature designs.

Yang \textit{et al.}~\cite{Yangshuai2021} decomposed face images into structure code and scalable color code for sparse encoding, then recovered images with the controllable adversarial generator for both machine and human vision. Similarly, Mao \textit{et al.}~\cite{Mao2023} proposed to disentangle face images into thumbnails and sketches for ultra-low bit-rate coding, while two-stage generative reconstructions with retrieval-guided external priors were leveraged to preserve face identity and reconstruction quality. To further utilize the strong generation ability of pretrained GAN model, a scalable GAN inversion was proposed for facial image decoding, where scalable latents were extracted by encoders for face attribute prediction and spatial feature transform of intermediate GAN features~\cite{Yagnwenhan2023}. With layer-wise hierarchical semantic information from StyleGANs~\cite{stylegan2}, Mao \textit{et al.}~\cite{Mao2024} utilized hierarchical style facial features from three-layer StyleGAN2 to realize reconstructions for coarse, middle, and high-level vision tasks. Apart from facial images, underwater image features could also realize extreme compression with the aid of an edge map, saliency mask, and foreground mask ~\cite{underwater1}. An enhancement layer then compressed the residual between machine-oriented reconstruction and original input for human-oriented reconstruction.

\subsection{Feature Domain Analysis}
Instead of performing machine analysis on traditional signal-level data, there has been a trend to bypass the reconstruction procedure and directly implement machine task on generative features. The feature could be intermediate features of generative codec or specially reconstructed features in parallel to reconstructed visual data. Similar to human-oriented scenarios, some of these works use single bit-stream with varied number of decoders for different tasks, while other works generate multiple bit-streams to represent scalable information of different granularity.

\subsubsection{Single-branch/Multi-branch Approaches}
Torfason \textit{et al.}~\cite{torfason2018towards} first explored the possibility of performing image understanding on the compressed representations by branching an inference network from latents in parallel with reconstruction decoder. Bai \textit{et al.}~\cite{bai2022towards} further extended this framework with transformer-based image analysis network and rate-distortion-accuracy optimization. To save computation for resource-constrained edge computing systems and knowledge distillation, Matsubara \textit{et al.}~\cite{matsubara2022supervised} proposed to discard visual reconstruction and directly reconstruct features from decoder for inference with a pruned task model. 

\subsubsection{Layered/Scalable Approaches}
To exploit the interaction between different levels of vision tasks and seek better representations for visual data, a scalable framework with multiple latents and bit-streams
~\cite{zhang2023machine} 
is more common for feature domain analysis. Choi \textit{et al.}~\cite{Choi2022} proposed to split latent representation into base layer and enhancement layer, where the former is transformed to the feature space for task back-end as well as reconstruction improvement. In addition, an adaptive partition-transmission-reconstruction-and-aggregation scheme~\cite{ICMH-Net} was proposed to select the optimal subset of latents of existing learning-based image codec for machine and human visions.

To further enhance the scalability of the whole network, Wang \textit{et al.}~\cite{shurun_face} proposed to use separately encoded face feature and texture as the base layer and enhancement layer. The analysis task is based on the base layer, while reconstruction is based on both base features and residuals from the enhancement layer. Similarly, content and style of face images are separately extracted and fused for multi-task learning ~\cite{DCC_multitask}. Such a scalable framework has also been successfully extended to video data with a two-layer structure 
~\cite{HMFVC}, where a semantic representation is learned to extract temporal semantic information and served for machine tasks as well as predictive coding for reconstruction. Moreover, a three-layer structure~\cite{SDVC} was further designed with semantic, structure and texture representations to satisfy different vision requirements. 

\section{Challenges and Research Directions}
Generative visual compression techniques have shown great potentials in compression efficiency and versatile vision analytics for natural image and video. Nevertheless, this does not imply that such generative visual compression research has been sufficiently developed and matured. The following drawbacks and challenges still exist and need to be solved,
\begin{itemize}
\item{ \textbf{Quality measures for evaluation and optimization}: Generative visual compression algorithms are designed in the feature domain, thus not being optimized for common pixel-level quality metrics in video coding, such as Peak Signal-to-Noise Ratio (PSNR) and Structural Similarity Index (SSIM). In addition, although existing perceptual-based quality measures like Deep Image Structure and Texture Similarity (DISTS)~\cite{dists} and Learned Perceptual Image Patch Similarity (LPIPS)~\cite{lpips} have been proposed for feature-domain measurement, they still lack some relevant priors and temporal learning for generative assessment. As a result, it is imperative to use appropriate perceptual assessment measures like~\cite{li2023perceptual} in order to precisely evaluate the performance effectiveness and optimize the algorithm design of generative compression tasks.}
\item{ \textbf{Robustness \& generalization capability}: Limited by the upper capabilities of deep generative models, the current reconstruction of generative compression algorithms is sometimes faced with some visual distortions and artifacts, resulting in an unpleasant visual quality of experience. Such instability in reconstruction quality damages the feasibility of practical applications to a certain extent. Therefore, it is important to explore more mature and robust technologies based on existing algorithms to improve the generalization ability of generative visual compression models.}
\item{ \textbf{Task-dependent compression \& communication}: These existing generative visual compression algorithms are mostly designed for specialized natural scenes, such as face and human body data with strong priors, so they are not task-independent and scenario-irrelevant. In real-scene applications, there is a high probability that visual data mixed with various objects/scenes will be encountered, which these existing algorithms cannot handle as well as traditional hybrid coding frameworks. As such, it is important to design a unified and universal generative compression algorithm for task-independent and scene-collaborated communication.}
\item{ \textbf{Standardization \& deployment}: In October 2023, the Joint Video Experts Team (JVET) of the ISO/IEC SC 29 and ITU-T SG16 has established a new Ad Hoc Group to conduct investigations~\cite{m64987,JVET-AG0042} on generative face video compression regarding software implementation, test conditions, coordinated experimentation, interoperability study, model lightweight and other aspects. Taking this as a starting point, it is hopeful that the standardization and deployment of other generative visual compression algorithms can be explored in the near future. In addition, to further optimize product design and user experience, including cost, performance, and power, hardware-software co-design of generative visual compression algorithm should be taken into account. }
\end{itemize}

\section{Conclusions}
This paper has made a comprehensive review on generative visual compression for both human and machine visions. Thanks to the strong inference capability of deep generative models, relevant visual compression for image and video data could be optimized and developed towards high coding efficiency, realistic signal reconstructions, and intelligent task analysis. As such, huge volumes of visual data around the world can be efficiently compressed via generative compression, whilst new applications and requirements in the post-AIGC era will further accelerate the research progress of generative visual compression techniques.

\bibliographystyle{IEEEbib}
\bibliography{refs}

\begin{thebibliography}{10}

\bibitem{8456298}
Shibani Santurkar et~al.,
\newblock ``Generative compression,''
\newblock in {\em Picture Coding Symposium}, 2018, pp. 258--262.

\bibitem{VAE}
Diederik~P Kingma and Max Welling,
\newblock ``Auto-encoding variational bayes,''
\newblock in {\em Proc. Int. Conf. Learn. Represent.}, 2014, p.~14.

\bibitem{goodfellow2014generative}
Ian Goodfellow et~al.,
\newblock ``Generative adversarial nets,''
\newblock {\em Proc. Adv. Neural Inf. Process. Syst.}, vol. 27, 2014.

\bibitem{ho2020denoising}
Jonathan Ho et~al.,
\newblock ``Denoising diffusion probabilistic models,''
\newblock {\em Proc. Adv. Neural Inf. Process. Syst.}, vol. 33, pp. 6840--6851, 2020.

\bibitem{radford2021learning}
Alec Radford et~al.,
\newblock ``Learning transferable visual models from natural language supervision,''
\newblock in {\em Proc. Int. Conf. Mach. Learn.}, 2021, pp. 8748--8763.

\bibitem{vaswani2017attention}
Ashish Vaswani et~al.,
\newblock ``Attention is all you need,''
\newblock {\em Proc. Adv. Neural Inf. Process. Syst.}, vol. 30, 2017.

\bibitem{wiegand2003overview}
Thomas Wiegand et~al.,
\newblock ``Overview of the {H.264/AVC} video coding standard,''
\newblock {\em IEEE Trans. Circuits Syst. Video Technol.}, vol. 13, no. 7, pp. 560--576, 2003.

\bibitem{sullivan2012overview}
Gary~J Sullivan et~al.,
\newblock ``Overview of the {H}igh {E}fficiency {V}ideo {C}oding ({HEVC}) standard,''
\newblock {\em IEEE Trans. Circuits Syst. Video Technol.}, vol. 22, no. 12, pp. 1649--1668, 2012.

\bibitem{bross2021overview}
Benjamin Bross et~al.,
\newblock ``Overview of the {Versatile Video Coding (VVC)} standard and its applications,''
\newblock {\em IEEE Trans. Circuits Syst. Video Technol.}, vol. 31, no. 10, 2021.

\bibitem{balle2016end}
Johannes Ball{\'e} et~al.,
\newblock ``End-to-end optimized image compression,''
\newblock in {\em Proc. Int. Conf. Learn. Represent.}, 2017.

\bibitem{balle2018variational}
Johannes Ball{\'e} et~al.,
\newblock ``Variational image compression with a scale hyperprior,''
\newblock in {\em Proc. Int. Conf. Learn. Represent.}, 2018.

\bibitem{NEURIPS2018_53edebc5}
David Minnen et~al.,
\newblock ``Joint autoregressive and hierarchical priors for learned image compression,''
\newblock in {\em Proc. Adv. Neural Inf. Process. Syst.}, 2018, vol.~31.

\bibitem{Cheng_2020_CVPR}
Zhengxue Cheng et~al.,
\newblock ``Learned image compression with discretized gaussian mixture likelihoods and attention modules,''
\newblock in {\em Proc. IEEE Conf. Comput. Vis. Pattern Recognit.}, June 2020.

\bibitem{minnen2020channel}
David Minnen et~al.,
\newblock ``Channel-wise autoregressive entropy models for learned image compression,''
\newblock in {\em Proc. IEEE Int. Conf. Image Process.} IEEE, 2020, pp. 3339--3343.

\bibitem{he2022elic}
Dailan He et~al.,
\newblock ``{ELIC}: Efficient learned image compression with unevenly grouped space-channel contextual adaptive coding,''
\newblock in {\em Proc. IEEE Conf. Comput. Vis. Pattern Recognit.}, 2022, pp. 5718--5727.

\bibitem{zhang2023elfic}
Zhichen Zhang et~al.,
\newblock ``{ELFIC}: A learning-based flexible image codec with rate-distortion-complexity optimization,''
\newblock in {\em Proc. ACM Int. Conf. Multimedia}, 2023, pp. 9252--9261.

\bibitem{yang2021slimmable}
Fei Yang et~al.,
\newblock ``Slimmable compressive autoencoders for practical neural image compression,''
\newblock in {\em Proceedings of the IEEE/CVF Conference on Computer Vision and Pattern Recognition}, 2021.

\bibitem{Agustsson_2019_ICCV}
Eirikur Agustsson et~al.,
\newblock ``Generative adversarial networks for extreme learned image compression,''
\newblock in {\em Proc. IEEE Int. Conf. Comput. Vis.}, October 2019.

\bibitem{NEURIPS2020_8a50bae2}
Fabian Mentzer et~al.,
\newblock ``High-fidelity generative image compression,''
\newblock in {\em Proc. Adv. Neural Inf. Process. Syst.}, 2020, vol.~33, pp. 11913--11924.

\bibitem{goose2023neural}
Noor~Fathima Goose et~al.,
\newblock ``Neural image compression with a diffusion-based decoder,''
\newblock {\em arXiv preprint arXiv:2301.05489}, 2023.

\bibitem{yang2022lossy}
Ruihan Yang and Stephan Mandt,
\newblock ``Lossy image compression with conditional diffusion models,''
\newblock {\em arXiv preprint arXiv:2209.06950}, 2022.

\bibitem{Habibian_2019_ICCV}
Amirhossein Habibian et~al.,
\newblock ``Video compression with rate-distortion autoencoders,''
\newblock in {\em Proc. IEEE Int. Conf. Comput. Vis.}, October 2019.

\bibitem{NEURIPS2019_f1ea154c}
Salvator Lombardo et~al.,
\newblock ``Deep generative video compression,''
\newblock in {\em Proc. Adv. Neural Inf. Process. Syst.}, 2019, vol.~32.

\bibitem{Li_2023_CVPR}
Jiahao Li et~al.,
\newblock ``Neural video compression with diverse contexts,''
\newblock in {\em Proc. IEEE Conf. Comput. Vis. Pattern Recognit.}, June 2023, pp. 22616--22626.

\bibitem{mentzer2021neural}
Fabian Mentzer et~al.,
\newblock ``Neural video compression using {GANs} for detail synthesis and propagation,''
\newblock in {\em Proc. Eur. Conf. Comput. Vis.}, 2022, pp. 562--578.

\bibitem{zhao_DPEG}
Tiesong Zhao et~al.,
\newblock ``Learning-based video coding with joint deep compression and enhancement,''
\newblock in {\em Proc. ACM Int. Conf. Multimedia}, 2022, p. 3045–3054.

\bibitem{Liu_2021_CVPR}
Bowen Liu et~al.,
\newblock ``Deep learning in latent space for video prediction and compression,''
\newblock in {\em Proc. IEEE Conf. Comput. Vis. Pattern Recognit.}, June 2021, pp. 701--710.

\bibitem{8269806}
Tadas Baltrušaitis et~al.,
\newblock ``Multimodal machine learning: A survey and taxonomy,''
\newblock {\em IEEE Trans. Pattern Anal. Mach. Intell.}, vol. 41, no. 2, pp. 423--443, 2019.

\bibitem{Liacmmm}
Jiguo Li et~al.,
\newblock ``Cross modal compression: Towards human-comprehensible semantic compression,''
\newblock in {\em Proc. ACM Int. Conf. Multimedia}, 2021, p. 4230–4238.

\bibitem{10032603}
Pingping Zhang et~al.,
\newblock ``Rethinking semantic image compression: Scalable representation with cross-modality transfer,''
\newblock {\em IEEE Trans. Circuits Syst. Video Technol.}, vol. 33, no. 8, pp. 4441--4445, 2023.

\bibitem{10125473}
Junlong Gao et~al.,
\newblock ``Rate-distortion optimization for cross modal compression,''
\newblock in {\em Data Compression Conference}, 2023, pp. 218--227.

\bibitem{10236483}
Junlong Gao et~al.,
\newblock ``Cross modal compression with variable rate prompt,''
\newblock {\em IEEE Trans. Multimedia}, pp. 1--13, 2023.

\bibitem{9953071}
Pulkit Tandon et~al.,
\newblock ``Txt2vid: Ultra-low bitrate compression of talking-head videos via text,''
\newblock {\em IEEE Journal on Selected Areas in Communications}, vol. 41, no. 1, pp. 107--118, 2023.

\bibitem{Lu_2022_CVPR}
Guo Lu et~al.,
\newblock ``Learning based multi-modality image and video compression,''
\newblock in {\em Proc. IEEE Conf. Comput. Vis. Pattern Recognit.}, June 2022, pp. 6083--6092.

\bibitem{jeon2014intrinsic}
Junho Jeon et~al.,
\newblock ``Intrinsic image decomposition using structure-texture separation and surface normals,''
\newblock in {\em European Conference Computer Vision}. Springer, 2014, pp. 218--233.

\bibitem{kim2018structure}
Youngjung Kim et~al.,
\newblock ``Structure-texture image decomposition using deep variational priors,''
\newblock {\em IEEE Trans. Image Process.}, vol. 28, no. 6, pp. 2692--2704, 2018.

\bibitem{chang2019layered}
Jianhui Chang et~al.,
\newblock ``Layered conceptual image compression via deep semantic synthesis,''
\newblock in {\em Proc. IEEE Int. Conf. Image Process.}, 2019, pp. 694--698.

\bibitem{chang2022conceptual}
Jianhui Chang et~al.,
\newblock ``Conceptual compression via deep structure and texture synthesis,''
\newblock {\em IEEE Trans. Image Process.}, vol. 31, pp. 2809--2823, 2022.

\bibitem{chang2022consistency}
Jianhui Chang et~al.,
\newblock ``Consistency-contrast learning for conceptual coding,''
\newblock in {\em Proc. ACM Int. Conf. Multimedia}, 2022, pp. 2681--2690.

\bibitem{chang2023semantic}
Jianhui Chang et~al.,
\newblock ``Semantic-aware visual decomposition for image coding,''
\newblock {\em Int. J. Comput. Vis.}, pp. 1--23, 2023.

\bibitem{chen2023generative}
Bolin Chen et~al.,
\newblock ``Generative face video coding techniques and standardization efforts: A review,''
\newblock {\em arXiv preprint arXiv:2311.02649}, 2023.

\bibitem{siarohin2019first}
Aliaksandr Siarohin et~al.,
\newblock ``First order motion model for image animation,''
\newblock {\em Proc. Adv. Neural Inf. Process. Syst.}, vol. 32, 2019.

\bibitem{facebook2021}
Maxime Oquab et~al.,
\newblock ``Low bandwidth video-chat compression using deep generative models,''
\newblock in {\em Proc. IEEE Conf. Comput. Vis. Pattern Recognit. Workshop}, 2021.

\bibitem{ultralow}
Goluck Konuko et~al.,
\newblock ``Ultra-low bitrate video conferencing using deep image animation,''
\newblock in {\em IEEE Int. Conf. Acoust. Speech Signal Process.}, 2021.

\bibitem{9859867}
Anni Tang et~al.,
\newblock ``Generative compression for face video: A hybrid scheme,''
\newblock in {\em Proc. IEEE Int. Conf. Multimedia Expo.} IEEE, 2022, pp. 1--6.

\bibitem{feng2021generative}
Dahu Feng et~al.,
\newblock ``A generative compression framework for low bandwidth video conference,''
\newblock in {\em Proc. IEEE Int. Conf. Multimedia Expo. Workshops}. IEEE, 2021.

\bibitem{chen2023interactive}
Bolin Chen et~al.,
\newblock ``Interactive face video coding: A generative compression framework,''
\newblock {\em arXiv preprint arXiv:2302.09919}, 2023.

\bibitem{CHEN2022DCC}
Bolin Chen et~al.,
\newblock ``Beyond keypoint coding: Temporal evolution inference with compact feature representation for talking face video compression,''
\newblock in {\em Data Compression Conference}, 2022.

\bibitem{10181664}
Ruofan Wang et~al.,
\newblock ``Extreme generative human-oriented video coding via motion representation compression,''
\newblock in {\em IEEE International Symposium on Circuits and Systems}, 2023, pp. 1--5.

\bibitem{9859831}
Ruofan Wang et~al.,
\newblock ``Disentangled visual representations for extreme human body video compression,''
\newblock in {\em Proc. IEEE Int. Conf. Multimedia Expo.}, 2022, pp. 1--6.

\bibitem{compressing2022bmvc}
Madhav Agarwal et~al.,
\newblock ``Compressing video calls using synthetic talking heads,''
\newblock in {\em British Machine Vision Conference}, 2023.

\bibitem{konuko2023predictive}
Goluck Konuko et~al.,
\newblock ``Predictive coding for animation-based video compression,''
\newblock in {\em Proc. IEEE Int. Conf. Image Process.}, 2023.

\bibitem{chen2023csvt}
Bolin Chen et~al.,
\newblock ``Compact temporal trajectory representation for talking face video compression,''
\newblock {\em IEEE Trans. Circuits Syst. Video Technol.}, 2023.

\bibitem{volokitin2022neural}
Anna Volokitin et~al.,
\newblock ``Neural face video compression using multiple views,''
\newblock in {\em Proc. IEEE Conf. Comput. Vis. Pattern Recognit.}, 2022.

\bibitem{icip2022zhao}
Zhao Wang et~al.,
\newblock ``Dynamic multi-reference generative prediction for face video compression,''
\newblock in {\em Proc. IEEE Int. Conf. Image Process.}, 2022, pp. 896--900.

\bibitem{yin2024parametertranslator}
Shanzhi Yin et~al.,
\newblock ``Enabling translatability of generative face video coding: A unified face feature transcoding framework,''
\newblock in {\em Data Compression Conference}, 2024.

\bibitem{li2023generative}
Zhengqi Li et~al.,
\newblock ``Generative image dynamics,''
\newblock {\em arXiv preprint arXiv:2309.07906}, 2023.

\bibitem{wang2021lossy}
Jianqiang Wang et~al.,
\newblock ``Lossy point cloud geometry compression via end-to-end learning,''
\newblock {\em IEEE Trans. Circuits Syst. Video Technol.}, vol. 31, no. 12, pp. 4909--4923, 2021.

\bibitem{he2022density}
Yun He et~al.,
\newblock ``Density-preserving deep point cloud compression,''
\newblock in {\em Proc. IEEE Conf. Comput. Vis. Pattern Recognit.}, 2022, pp. 2333--2342.

\bibitem{nguyen2021lossless}
Dat~Thanh Nguyen et~al.,
\newblock ``Lossless coding of point cloud geometry using a deep generative model,''
\newblock {\em IEEE Trans. Circuits Syst. Video Technol.}, vol. 31, no. 12, pp. 4617--4629, 2021.

\bibitem{jia2018light}
Chuanmin Jia et~al.,
\newblock ``Light field image compression using generative adversarial network-based view synthesis,''
\newblock {\em IEEE Trans. Circuits Syst. Video Technol.}, vol. 9, no. 1, pp. 177--189, 2018.

\bibitem{liu2021view}
Deyang Liu et~al.,
\newblock ``View synthesis-based light field image compression using a generative adversarial network,''
\newblock {\em Information Sciences}, vol. 545, pp. 118--131, 2021.

\bibitem{jiang2022untrained}
Xiaoran Jiang et~al.,
\newblock ``An untrained neural network prior for light field compression,''
\newblock {\em IEEE Trans. Image Process.}, vol. 31, 2022.

\bibitem{bakir2020light}
Nader Bakir et~al.,
\newblock ``Light field image coding using dual discriminator generative adversarial network and vvc temporal scalability,''
\newblock in {\em Proc. IEEE Int. Conf. Multimedia Expo.} IEEE, 2020, pp. 1--6.

\bibitem{liu2019dsic}
Jerry Liu et~al.,
\newblock ``Dsic: Deep stereo image compression,''
\newblock in {\em Proc. IEEE Int. Conf. Comput. Vis.}, 2019, pp. 3136--3145.

\bibitem{li2022end}
Mu~Li et~al.,
\newblock ``End-to-end optimized 360° image compression,''
\newblock {\em IEEE Trans. Image Process.}, vol. 31, pp. 6267--6281, 2022.

\bibitem{Duan_2022}
Lingyu Duan et~al.,
\newblock ``Video coding for machines: A paradigm of collaborative compression and intelligent analytics,''
\newblock {\em IEEE Trans. Image Process.}, vol. 29, pp. 8680--8695, 2020.

\bibitem{e2e4m_study_DCC2022}
Lahiru~D. Chamain et~al.,
\newblock ``End-to-end optimized image compression for machines, a study,''
\newblock in {\em Data Compression Conference}, 2021.

\bibitem{Le_weighting}
Nam Le et~al.,
\newblock ``Image coding for machines: an end-to-end learned approach,''
\newblock in {\em IEEE Int. Conf. Acoust. Speech Signal Process.}, 2021, pp. 1590--1594.

\bibitem{Le_infertune}
Nam Le et~al.,
\newblock ``Learned image coding for machines: A content-adaptive approach,''
\newblock in {\em Proc. IEEE Int. Conf. Multimedia Expo.}, 2021.

\bibitem{wang2022end}
Shurun Wang et~al.,
\newblock ``Deep image compression toward machine vision: A unified optimization framework,''
\newblock {\em IEEE Trans. Circuits Syst. Video Technol.}, vol. 33, no. 6, pp. 2979--2989, 2023.

\bibitem{wang2021end}
Shurun Wang et~al.,
\newblock ``End-to-end compression towards machine vision: Network architecture design and optimization,''
\newblock {\em IEEE open j. circuits syst.}, vol. 2, pp. 675--685, 2021.

\bibitem{underwater2}
Zhengkai Fang et~al.,
\newblock ``Prior-guided contrastive image compression for underwater machine vision,''
\newblock {\em IEEE Trans. Circuits Syst. Video Technol.}, vol. 33, no. 6, pp. 2950--2961, 2023.

\bibitem{Yi2023}
Xiaokai Yi et~al.,
\newblock ``Task-driven video compression for humans and machines: Framework design and optimization,''
\newblock {\em IEEE Trans. Multimedia}, vol. 25, pp. 8091--8102, 2023.

\bibitem{chamain2021end}
Lahiru~D Chamain et~al.,
\newblock ``End-to-end optimized image compression for multiple machine tasks,''
\newblock {\em arXiv preprint arXiv:2103.04178}, 2021.

\bibitem{TransIC2023}
Yi-Hsin Chen et~al.,
\newblock ``Transtic: Transferring transformer-based image compression from human perception to machine perception,''
\newblock in {\em Proc. IEEE Int. Conf. Comput. Vis.}, 2023, pp. 23240--23250.

\bibitem{Yangshuai2021}
Shuai Yang et~al.,
\newblock ``Towards coding for human and machine vision: Scalable face image coding,''
\newblock {\em IEEE Trans. Multimedia}, vol. 23, pp. 2957--2971, 2021.

\bibitem{Mao2023}
Yudong Mao et~al.,
\newblock ``Peering into the sketch: Ultra-low bitrate face compression for joint human and machine perception,''
\newblock in {\em Proc. ACM Int. Conf. Multimedia}, 2023, p. 2564–2572.

\bibitem{Yagnwenhan2023}
Wenhan Yang et~al.,
\newblock ``Facial image compression via neural image manifold compression,''
\newblock {\em IEEE Trans. Circuits Syst. Video Technol.}, pp. 1--1, 2023.

\bibitem{stylegan2}
Tero Karras et~al.,
\newblock ``Analyzing and improving the image quality of stylegan,''
\newblock in {\em Proc. IEEE Conf. Comput. Vis. Pattern Recognit.}, 2020.

\bibitem{Mao2024}
Qi~Mao et~al.,
\newblock ``Scalable face image coding via stylegan prior: Toward compression for human-machine collaborative vision,''
\newblock {\em IEEE Trans. Image Process.}, vol. 33, pp. 408--422, 2024.

\bibitem{underwater1}
Zhengkai Fang et~al.,
\newblock ``Priors guided extreme underwater image compression for machine vision and human vision,''
\newblock {\em IEEE J. Ocean. Eng.}, vol. 48, no. 3, pp. 888--902, 2023.

\bibitem{torfason2018towards}
Robert Torfason et~al.,
\newblock ``Towards image understanding from deep compression without decoding,''
\newblock {\em arXiv preprint arXiv:1803.06131}, 2018.

\bibitem{bai2022towards}
Yuanchao Bai et~al.,
\newblock ``Towards end-to-end image compression and analysis with transformers,''
\newblock in {\em Proc. AAAI Conf. Artif. Intell.}, 2022.

\bibitem{matsubara2022supervised}
Yoshitomo Matsubara et~al.,
\newblock ``Supervised compression for resource-constrained edge computing systems,''
\newblock in {\em Proceedings of the IEEE/CVF Winter Conference on Applications of Computer Vision}, 2022, pp. 2685--2695.

\bibitem{zhang2023machine}
Yuefeng Zhang et~al.,
\newblock ``Machine perception-driven image compression: A layered generative approach,''
\newblock {\em arXiv preprint arXiv:2304.06896}, 2023.

\bibitem{Choi2022}
Hyomin Choi et~al.,
\newblock ``Scalable image coding for humans and machines,''
\newblock {\em IEEE Trans. Image Process.}, vol. 31, pp. 2739--2754, 2022.

\bibitem{ICMH-Net}
Lei Liu et~al.,
\newblock ``Icmh-net: Neural image compression towards both machine vision and human vision,''
\newblock in {\em Proc. ACM Int. Conf. Multimedia}, 2023, p. 8047–8056.

\bibitem{shurun_face}
Shurun Wang et~al.,
\newblock ``Towards analysis-friendly face representation with scalable feature and texture compression,''
\newblock {\em IEEE Trans. Multimedia}, vol. 24, pp. 3169--3181, 2022.

\bibitem{DCC_multitask}
Yuefeng Zhang et~al.,
\newblock ``Analysis on compressed domain: A multi-task learning approach,''
\newblock in {\em Data Compression Conference}, 2022, p. 494.

\bibitem{HMFVC}
Zhimeng Huang et~al.,
\newblock ``Hmfvc: A human-machine friendly video compression scheme,''
\newblock {\em IEEE Trans. Circuits Syst. Video Technol.}, pp. 1--1, 2022.

\bibitem{SDVC}
Hongbin Lin et~al.,
\newblock ``{DeepSVC}: Deep scalable video coding for both machine and human vision,''
\newblock in {\em Proc. ACM Int. Conf. Multimedia}, 2023, p. 9205–9214.

\bibitem{dists}
Keyan Ding et~al.,
\newblock ``Image quality assessment: Unifying structure and texture similarity,''
\newblock {\em IEEE Trans. Pattern Anal. Mach. Intell.}, 2020.

\bibitem{lpips}
Richard Zhang et~al.,
\newblock ``The unreasonable effectiveness of deep features as a perceptual metric,''
\newblock in {\em Proc. IEEE Conf. Comput. Vis. Pattern Recog.}, 2018.

\bibitem{li2023perceptual}
Yixuan Li et~al.,
\newblock ``Perceptual quality assessment of face video compression: A benchmark and an effective method,''
\newblock {\em arXiv preprint arXiv:2304.07056}, 2023.

\bibitem{m64987}
Yan Ye et~al.,
\newblock ``On {VVC}-assisted ultra-low rate generative face video coding,''
\newblock {\em {MPEG ISO/IEC JTC 1/SC 29/WG 2 doc. no. m64987}}, October 2023.

\bibitem{JVET-AG0042}
Bolin Chen et~al.,
\newblock ``{AHG16}: Proposed common software tools and testing conditions for generative face video compression,''
\newblock {\em {The Joint Video Experts Team of ITU-T SG 16 WP 3 and ISO/IEC JTC 1/SC 29, doc. no. JVET-AG0042}}, January 2024.

\end{thebibliography}
\end{document}